\documentclass[letterpaper, 10 pt, conference]{ieeeconf}  

\IEEEoverridecommandlockouts                              

\usepackage{url}
\usepackage{amsmath}
\usepackage{graphicx}
\usepackage{verbatim}
\usepackage{algorithm}
\usepackage{algorithmic}
\usepackage{hyperref}
\usepackage{color}
\usepackage{wrapfig}
\usepackage{array}
\usepackage{amssymb}
\usepackage[small]{caption}

\long\def\ignorethis#1{}

\newcommand{\nth}{\text{th}}
\newcommand{\pr}{^\prime}
\newcommand{\tr}{^\mathrm{T}}
\newcommand{\inv}{^{-1}}

\newcommand{\gauss}{\mathcal{N}}

\newcommand{\trace}{\text{tr}}

\newcommand{\trajdist}{p}
\newcommand{\policy}{\pi}

\newcommand{\params}{\theta}

\newcommand{\cost}{\ell}
\newcommand{\state}{\mathbf{x}}
\newcommand{\action}{\mathbf{u}}
\newcommand{\obs}{\mathbf{o}}
\newcommand{\hstate}{\hat{\mathbf{x}}}
\newcommand{\haction}{\hat{\mathbf{u}}}

\newcommand{\traj}{\tau}

\newcommand{\polmu}{\mu^\policy}
\newcommand{\polsig}{\Sigma^\policy}

\newcommand{\detpolicy}{g}

\newcommand{\bu}{\mathbf{u}}

\newcommand{\kl}{D_\text{KL}}
\newcommand{\ent}{\mathcal{H}}

\newcommand{\fct}{f_{c t}}
\newcommand{\fxt}{f_{\state t}}
\newcommand{\fut}{f_{\action t}}
\newcommand{\fxtpr}{f_{\state t\pr}}
\newcommand{\futpr}{f_{\action t\pr}}

\newcommand{\fyt}{f_{\state\action t}}

\newcommand{\Qxt}{Q_{\state t}}
\newcommand{\Qut}{Q_{\action t}}
\newcommand{\Qyt}{Q_{\state\action t}}
\newcommand{\Qxxt}{Q_{\state,\state t}}
\newcommand{\Quut}{Q_{\action,\action t}}

\newcommand{\tQuutij}{\tilde{Q}_{\action,\action t i j}}
\newcommand{\Quutpr}{Q_{\action,\action t\pr}}
\newcommand{\Quxt}{Q_{\action,\state t}}

\newcommand{\Qyyt}{Q_{\state\action,\state\action t}}

\newcommand{\Vxt}{V_{\state t}}
\newcommand{\Vxxt}{V_{\state,\state t}}
\newcommand{\Vxtp}{V_{\state t+1}}
\newcommand{\Vxxtp}{V_{\state,\state t+1}}
\newcommand{\kpol}{\mathbf{k}}
\newcommand{\Kpol}{\mathbf{K}}

\newcommand{\noise}{\mathbf{F}}

\newcommand{\costgradt}{\cost_{\state\action t}}
\newcommand{\costhesst}{\cost_{\state\action,\state\action t}}
\newcommand{\tcostgradt}{\tilde{\cost}_{\state\action t}}
\newcommand{\tcosthesst}{\tilde{\cost}_{\state\action,\state\action t}}

\newcommand{\st}{\state_t}
\newcommand{\at}{\action_t}
\newcommand{\ot}{\obs_t}

\newcommand{\stv}{\mathbf{v}_t}
\newcommand{\stq}{\mathbf{q}_t}
\newcommand{\stp}{\mathbf{p}_t}
\newcommand{\stw}{\boldsymbol{\omega}_t}
\newcommand{\str}{\mathbf{r}_t}

\newcommand{\lgmut}{\lambda_{\mu t}}
\newcommand{\lgmutpr}{\lambda_{\mu t\pr}}
\newcommand{\admmrho}{\nu}

\newcolumntype{C}[1]{>{\centering\let\newline\\\arraybackslash\hspace{0pt}}m{#1}}

\title{\LARGE \bf
Learning Deep Control Policies for Autonomous Aerial Vehicles with MPC-Guided Policy Search
}

\author{Tianhao Zhang, Gregory Kahn, Sergey Levine, Pieter Abbeel
\thanks{Department of Electrical Engineering and Computer Science, University of California, Berkeley, Berkeley, CA 94720}%
}

\begin{document}

\maketitle
\thispagestyle{empty}
\pagestyle{empty}

\begin{abstract}

Model predictive control (MPC) is an effective method for controlling robotic systems, particularly autonomous aerial vehicles such as quadcopters. However, application of MPC can be computationally demanding, and typically requires estimating the state of the system, which can be challenging in complex, unstructured environments. Reinforcement learning can in principle forego the need for explicit state estimation and acquire a policy that directly maps sensor readings to actions, but is difficult to apply to unstable systems that are liable to fail catastrophically during training before an effective policy has been found. We propose to combine MPC with reinforcement learning in the framework of guided policy search, where MPC is used to generate data at training time, under full state observations provided by an instrumented training environment. This data is used to train a deep neural network policy, which is allowed to access only the raw observations from the vehicle's onboard sensors. After training, the neural network policy can successfully control the robot without knowledge of the full state, and at a fraction of the computational cost of MPC. We evaluate our method by learning obstacle avoidance policies for a simulated quadrotor, using simulated onboard sensors and no explicit state estimation at test time.


\end{abstract}

\section{Introduction}

Model predictive control (MPC) is an effective and reliable method for controlling robotic systems, particularly autonomous aerial vehicles such as quadcopters, because of its robustness to moderate model errors \cite{msr-rmpcc-05}, ability to use high-level objectives \cite{tet-mjc-12}, and relative simplicity. However, applications of MPC can be computationally demanding, and typically require estimating the state of the system. The state estimation problem can be quite challenging in complex, unstructured environments. Reinforcement learning can in principle forego the need for explicit state estimation and acquire a policy that directly maps sensor readings to actions \cite{dnp-spsr-13}. The power of reinforcement learning is derived from its ability to learn directly from the real-world behavior of the system. Unfortunately, this strength is also its major weakness when applied to unstable, fragile systems such as aerial vehicles, which can be damaged beyond repair by an unsuccessful, partially trained policy (e.g. by crashing into an obstacle). While alternative learning methods, such as learning from demonstration \cite{rgb-rilsp-11,rmswd-lmruc-13}, can address this issue, they typically require costly additional information, such as guidance from a human expert.

We propose to use an off-policy guided policy search algorithm in combination with a model predictive control (MPC) scheme to train policies for autonomous aerial vehicles in a way that avoids catastrophic failure at training time. Guided policy search transforms RL into supervised learning, where an optimal control algorithm provides the supervision and the final control policy is trained with supervised learning. Typically, this optimal control algorithm is an offline trajectory optimization procedure, which either assumes a known model of the dynamics \cite{lk-lcnnp-14} or uses an iteratively learned model \cite{la-lnnpg-14}. Both approaches are prone to failure during training because the known model may be inaccurate while the learned model is always inaccurate during the early stages of learning. By substituting MPC for offline trajectory optimization, we can obtain a variant of guided policy search that is robust to moderate model errors, and thus avoid catastrophic failures during training. Furthermore, since the final policy is trained with supervised learning, we can train complex, high-dimensional, and highly nonlinear policies, such as deep neural networks, which can represent a wide range of complex behaviors.

\begin{figure}
\setlength{\unitlength}{0.5\columnwidth}
\includegraphics[width=\columnwidth]{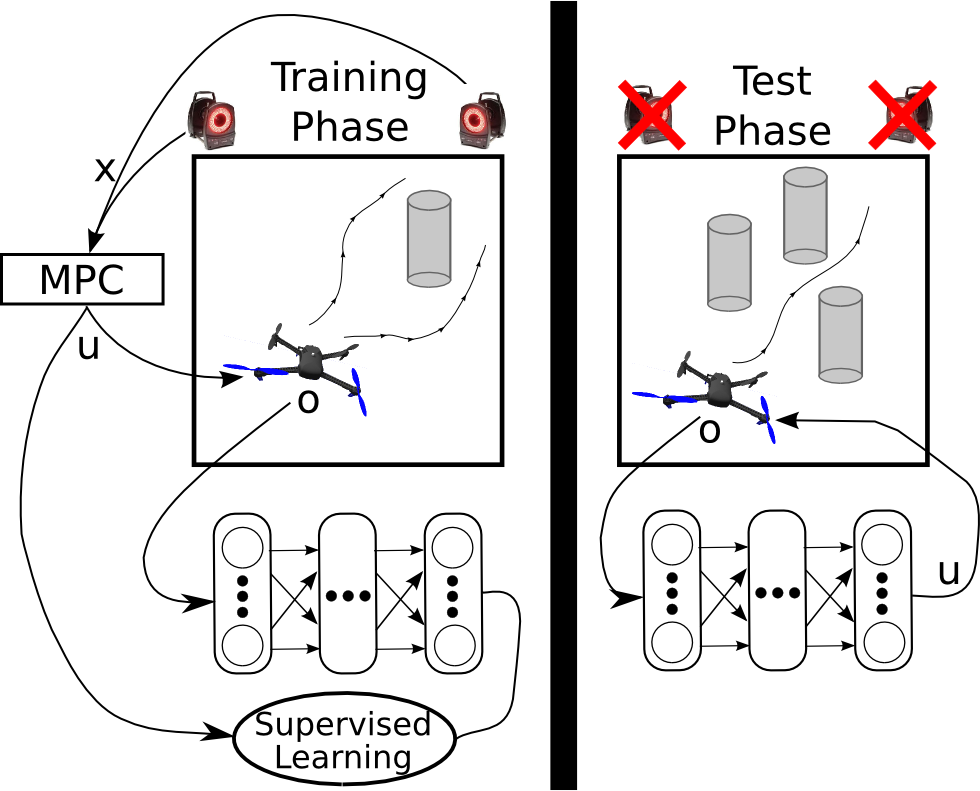}
\caption{Diagram of our method: the training phase alternates between running MPC to attempt the task and collect data under full state observation, and using this data to train a neural network policy that chooses actions based only on the vehicle's onboard sensors. At test time, the neural network does not need the full state, and can control the vehicle in unstructured environments.
\label{fig:teaser}
}
\vspace{-0.2in}
\end{figure}

One might wonder why the guided policy search method is necessary if we already have access to an effective MPC procedure. In the case of autonomous aerial vehicles, training deep neural network policies with guided policy search affords us two main advantages. First, the neural network policy does not need to use the same inputs as MPC. In fact, we can restrict its inputs to only those observations that are directly available from the vehicle's onboard sensors, such as IMU readings and data from laser range finders, while the MPC training phase uses the full state of the system. Since the policy is represented by a deep neural network, it can even process complex, raw sensor information. For example, prior work has shown that guided policy search can learn policies that directly use camera images \cite{levine2015end}. Since MPC is only used at training time, we can employ an instrumented training setup, where the full state is known at training time (e.g. using motion capture), but unavailable at test time. This instrumented training setup is one of the key benefits of our approach, since it allows for safe training with full state information, but still produces a policy that uses raw sensor readings and does not require explicit state estimation. The second benefit of this approach is that the final neural network policy is computationally much less expensive than MPC, and can be easily parallelized on specialized hardware. This advantage combines elegantly with the instrumented training setup, since the MPC solution can be computed offboard during training, while all policy computations may be performed onboard at test time.






Our main contribution is an MPC-guided policy search algorithm that can be used for learning control policies for autonomous aerial vehicles. This algorithm, illustrated in Figure~\ref{fig:teaser}, replaces the offline trajectory optimization that is typically used in guided policy search with online MPC, which continuously replans paths to the goal from the vehicle's current state using an approximate model of the dynamics. Our modified MPC procedure also takes into account the actions that would be taken at each state by the current neural network in order to avoid actions that the network is unlikely to take. This ensures that, at convergence, the neural network achieves good long-horizon performance, despite being trained only with supervised learning. Our approach allows us to learn neural network policies that directly process raw observations from the vehicle's onboard sensors, and are substantially faster to evaluate at test time than full MPC solutions. We demonstrate our method on a set of simulated quadrotor control tasks, including obstacle avoidance using simulated laser range sensors. We show that our approach can learn policies that are robust to a variety of perturbations and generalize successfully to large obstacle courses, without catastrophic failure during training.



\section{Related Work}

Model predictive control (MPC) is an effective and popular technique for control of robotic systems, and is frequently used to control autonomous aerial vehicles such as quadrotors \cite{sks-nmptc-02,r-rcmpc-04,ror-ipnc-10,asd-gcftq-12,ant-mpqca-12,bat-lbmpc-12,md-mpcqs-13}. MPC is straightforward to apply when the state of the system is known (e.g. via a motion capture system), or when it can be measured accurately through sensors with well-understood observation models. However, vehicles navigating complex, unstructured environments must use more complex sensors, such as cameras and laser range finders. Incorporating such sensors into optimal control directly is challenging, since the sensor reading depends on a complex and often unknown environment. This challenge is conventionally addressed by using localization and mapping algorithms to map out the environment \cite{fhhlm-vbame-12} and then optimizing trajectories under the resulting map \cite{hmtfp-aoamv-11}. However, this kind of model-based approach is quite challenging when the vehicle is moving at high speed, or when onboard computation is limited.

On the other end of the spectrum from such model-based methods, reinforcement learning (RL) aims to directly learn control policies that map observations to controls \cite{dnp-spsr-13}. This approach in principle removes the need for explicit state estimation and extensive computation at test time, by using a number of training episodes to iteratively improve the policy from real-world experience. RL has been used to train robotic controllers for games such as ball-in-cup and table tennis \cite{kop-rlarm-10}, manipulation \cite{phas-lgmsl-09,drf-lclcm-11}, and robotic locomotion \cite{kp-pgrlf-04,tzs-spgrl-04,gpw-fbwrc-06,emmnc-lcbbl-08}. An overview of recent reinforcement learning methods in robotics can be found in a recent survey paper \cite{kbp-rlrs-13}. However, model-free RL is difficult to apply to unstable systems such as quadrotors, due to the possibility of catastrophic failure during training. Model-based RL can mitigate this problem by training a model from real-world experience, and then optimizing the policy under this model. While such methods have been successfully applied to aerial vehicles \cite{acqn-arlah-06,mueller-iros12}, the requirement to be able to acquire an accurate model means that these methods share many of the challenges of MPC methods.

In this work, we use an off-policy reinforcement learning method called guided policy search, which incorporates the advantages of model-based methods at training time, while still training the policy to use only the onboard sensors of the robot, without explicit state estimation and using only real-world data. Guided policy search has been applied to locomotion \cite{lk-lcnnp-14}, robotic manipulation \cite{lwa-lnnpg-15}, and vision-based robotic control \cite{levine2015end}, but all prior applications rely on an offline trajectory optimization phase to generate the controller that is then executed on the real system. While this offline optimization might use a learned model of the system dynamics \cite{la-lnnpg-14}, the resulting trajectory-centric controller is only adapted between episodes. This makes these methods liable to fail catastrophically when the model is inaccurate. We replace the offline trajectory optimization in guided policy search with MPC, which prevents catastrophic failures even during training, making the method suitable for learning policies for autonomous aerial vehicles.

One of the key advantages of guided policy search is its ability to train complex, high-dimensional, and highly nonlinear policies \cite{lk-gps-13}. This allows us to use deep neural network representations for our policies, thus allowing them to handle complex, raw input from onboard sensors, without extensive engineering of the policy parameterization. While neural networks have been used for control for decades \cite{hszg-nncss-92,p-alvin-89}, limitations on computation and algorithms have made large neural network policies very difficult to learn. More recently, deep neural network policies have been used for tasks ranging from robotic control \cite{lwa-lnnpg-15,levine2015end} to video game playing \cite{mksga-padrl-13}. Deep neural networks have also been used to learn models for MPC \cite{lks-dmpc-15}. In this paper, we use neural networks to represent the policy, rather than the model, while MPC is used to help train this policy.

\section{Preliminaries}

The goal of policy search is to minimize the expected cost $E_{\policy_\params}\big[\sum_{t=1}^T \cost(\st,\at)\big]$ with respect to the parameters $\params$ of a policy $\policy_\params(\at|\ot)$. Here, $\st$ denotes the full state at time $t$, $\ot$ the observation, $\at$ the action, and $\cost(\st,\at)$ the cost function that defines the task. For example, a task that requires a quadrotor to fly to a position might have the cost be the distance between the vehicle and the goal. The expectation is taken with respect to \mbox{$p(\traj) = p(\state_1)\prod_{t=1}^T p(\state_{t+1}|\st,\at) \policy_\params(\at|\st)$}, where \mbox{$\traj = \{\state_1,\action_1,\dots,\state_T,\action_T\}$} denotes a trajectory. The observation $\ot$ is distributed according to some unknown observation distribution $p(\ot|\st)$, which describes how the readings from the robot's sensors depend on the state.

Optimal control also seeks to solve problems of this type, under various assumptions about the dynamics $p(\state_{t+1}|\st,\at)$, observation function $p(\ot|\st)$, and policy. For example, the differential dynamic programming (DDP) algorithm can be viewed as approximately optimizing the expected cost under a locally linear-Gaussian dynamics model and with time-varying linear \cite{jm-ddp-70,lt-ilqr-04} or linear-Gaussian \cite{lk-gps-13} policies that act directly on the state (i.e. $\ot = \st$). While this method is not as general as policy gradient RL algorithms, which can optimize arbitrary parameterized policies with arbitrary observations under unknown dynamics \cite{ps-rlmsp-08}, it is fast, simple, and often effective \cite{tet-sscbo-12}. In order to combine the efficiency of DDP with the flexibility of general policy search methods, guided policy search uses DDP-like algorithms to solve the control problem from a variety of initial states and generate training data for arbitrary parameterized policies, which are then trained with supervised learning to mimic the behavior of the DDP solutions \cite{lk-gps-13,la-lnnpg-14}.

\begin{algorithm}[b]
\caption{Generic guided policy search summary}
\label{alg:gps}
\begin{algorithmic}[1]
\FOR{iteration $k=1$ to $K$}
\STATE Optimize trajectory distributions $\trajdist_i(\traj)$ to minimize $E_{\trajdist_i}[\cost(\traj)]$ and deviation from the policy $\policy_\params(\at|\ot)$
\STATE Generate samples $\{\traj_i^j\}$ from each $\trajdist_i(\traj)$
\STATE Train nonlinear policy $\policy_\params(\at|\ot)$ to match the sampled trajectories $\{\traj_i^j\}$
\STATE Update Lagrange multipliers to encourage agreement between $\trajdist_i(\at|\st)$ and $\policy_\params(\at|\st)$
\ENDFOR
\STATE {\bf return} optimized policy parameters $\params$
\end{algorithmic}
\end{algorithm}

Algorithm~\ref{alg:gps} presents a generic guided policy search method, where trajectory optimization is used to optimize a set of guiding trajectory distributions $\trajdist_i(\traj)$, defined by the corresponding linear-Gaussian controllers $\trajdist_i(\at|\st)$, and an arbitrary nonlinear policy $\policy_\params(\at|\ot)$ is trained using samples from all of these controllers. Since supervised learning does not in general guarantee that the policy $\policy_\params(\at|\ot)$ will achieve good long-horizon performance~\cite{rgb-rilsp-11}, guided policy search alternates between optimizing the policy and optimizing each of the trajectory distributions, each time adjusting the trajectory cost and the policy optimization objective to ensure that the linear-Gaussian controllers $\trajdist_i(\at|\st)$ and policy $\policy_\params(\at|\ot)$ converge to the same behavior. The objective for trajectory optimization is modified by adding a penalty for the deviation from the policy, and the policy objective is modified by applying different weights to different samples \cite{la-lnnpg-14} or using dual variables \cite{levine2015end}. Convergence to a policy $\policy_\params(\at|\ot)$ that minimizes expected cost can be shown by casting this alternating optimization as a relaxation of a constrained optimization problem of the form
\begin{align*}
\min_{\params,\trajdist} \, & E_{\traj\sim\trajdist}[\cost(\traj)] \\
\text{s.t. } & \trajdist(\at|\st) = \policy_\params(\at|\st) \,\, \forall t
\end{align*}
\noindent where $\cost(\traj)$ is shorthand for $\sum_{t=1}^T \cost(\st,\at)$, $\policy_\params(\at|\st)$ is shorthand for $\int \policy_\params(\at|\ot)p(\ot|\st)d\ot$, and $\trajdist(\traj)$ is a mixture of the guiding distributions $\trajdist_i(\traj)$. This constrained problem can be approximately transformed into the alternating optimization in Algorithm~\ref{alg:gps} by using sampling over the observations $\ot$ together with the framework of dual gradient descent \cite{la-lnnpg-14} or BADMM \cite{levine2015end}. In this paper, we use the BADMM version, which specifies the following objective for trajectory optimization:
\begin{small}
\begin{multline}
\min_{\trajdist_i(\traj)} \sum_{t=1}^T  E_{\trajdist_i(\st,\at)}\Big[ \cost(\st,\at) - \at\tr\lgmut^i \\
+ \admmrho_t^i \kl\big(\trajdist_i(\at|\st)\|\policy_\params(\at|\st)\big) \Big] \label{eqn:trajobj}
\end{multline}
\end{small}%
where $\lgmut$ is a Lagrange multiplier on the mean action, and the third term is a KL-divergence penalty. Together, these terms serve to keep $\trajdist_i(\at|\st)$ close to $\policy_\params(\at|\st)$. The supervised objective for the policy is similarly given by
\begin{small}
\begin{multline}
\min_\params \sum_{i=1}^N \sum_{j=1}^M \sum_{t=1}^T \Big[\admmrho_t^i \kl\big(\policy_\params(\at|\ot^{i,j})\|\trajdist_i(\at|\st^{i,j})\big)\\
+ E_{\policy_\params(\at|\ot^{i,j})}[\at]\tr \lgmut^i\Big] \label{eqn:polobj}
\end{multline}
\end{small}%
where $N$ is the number of trajectory distributions $\trajdist_i$, $M$ is the number of samples collected from each $\trajdist_i(\traj)$ and $T$ is the length of each trajectory $\traj$. This objective uses samples to estimate the integral $\int \policy_\params(\at|\ot)p(\ot|\st)d\ot$ and, in the case where the policy is given by the Gaussian $\gauss\big(\polmu(\ot),\polsig\big)$, it corresponds to a weighted least squares objective on the mean $\polmu(\ot)$, while $\polsig$ can be solved for in closed form. For a detailed derivation of this method, as well as the update equations for $\admmrho_t^i$ and $\lgmut^i$, we refer the reader to previous work \cite{levine2015end}. Note that the policy $\policy_\params(\at|\ot)$ only uses the observations $\ot$ as input, which means that, once it has been trained, it can be used in situations where the true state $\st$ is unknown.

Prior guided policy search methods optimized the guiding trajectory distributions $\trajdist_i(\traj)$ using either offline trajectory optimization with known system dynamics \cite{lk-lcnnp-14}, or trajectory-centric reinforcement learning \cite{la-lnnpg-14}. The former class of methods assumes that the true dynamics are known in advance, while the latter requires iteratively learning the dynamics by attempting to run potentially suboptimal controllers on the real physical system. In the case of unstable systems, such as autonomous aerial vehicles, neither approach is ideal, since the true dynamics are not known perfectly, and the suboptimal controller rollouts required for reinforcement learning might cause catastrophic failure, such as a crash. On the other hand, MPC methods that continuously recompute the vehicle's trajectory under an approximate model of the dynamics have been shown to exhibit good robustness to model errors \cite{tet-sscbo-12}. In the next section, we discuss how MPC can be combined with guided policy search to learn effective control policies.

\section{MPC-Guided Policy Search}

In this paper, we use MPC together with offline trajectory optimization to generate guiding samples for guided policy search. We assume that we have access to an approximate model of the system dynamics, which we use during training to choose actions that will accomplish the desired task, starting from a variety of initial states. These samples are then used as training data to train a nonlinear policy $\policy_\params(\at|\ot)$, and this policy is included in the cost function for the next batch of samples. By repeatedly collecting new samples and training the policy $\policy_\params(\at|\ot)$ in this way, the method can acquire an effective nonlinear policy that generalizes to new states.

This method is a special case of the generic guided policy search framework presented in Algorithm~\ref{alg:gps}, but a number of modifications are necessary to adapt the approach to use MPC to generate the guiding trajectory distributions. First, the MPC procedure must minimize the objective in Equation~(\ref{eqn:trajobj}), which means that it must also minimize deviation from the neural network policy. Second, each MPC rollout produces a different locally linear controller, which necessitates a modification to the supervised policy learning phase. Lastly, since MPC uses a relatively short horizon, we generate target trajectories using an offline trajectory optimization phase, and then track these trajectories. We develop a formulation for this tracking objective that is compatible with guided policy search.

\subsection{Model Predictive Control with DDP}

The MPC method we use is based on differential dynamic programming (DDP) \cite{jm-ddp-70}. In particular, we use an efficient variant of this method called iterative LQG, which assumes access to an approximate model of the system dynamics and uses a local linear-quadratic expansion to solve the optimal control problem. We summarize the method in this section. However, our derivation largely follows prior work \cite{tet-mjc-12}.

Iterative LQG assumes that the dynamics are given by a deterministic mean function \mbox{$f(\st,\at) = E[\state_{t+1}|\st,\at]$}, with additive Gaussian noise. The algorithm computes a linear expansion of the dynamics around a nominal trajectory \mbox{$\hat{\traj} = \{\hat{\state}_1,\hat{\action}_1,\dots,\hat{\state}_T,\hat{\action}_T\}$}, as well as a quadratic expansion of the cost. Without loss of generality, we assume that the nominal states and actions are zero for notational convenience. The linearized dynamics have the form $p(\state_{t+1}|\st,\at) = \gauss(\fxt\st + \fut\at + \fct, \noise_t)$, and the quadratic cost approximation has the form
\[
\cost(\st,\at) \approx \frac{1}{2}[\st;\at]\tr\costhesst[\st;\at] + [\st;\at]\tr\costgradt + \text{const},
\]
\noindent where subscripts denote derivatives, e.g. $\tcostgradt$ is the gradient of $\tilde{\cost}$ with respect to $[\st;\at]$, while $\tcosthesst$ is the Hessian. 
Under this model of the dynamics and cost function, the optimal policy can be computed by recursively computing the quadratic $Q$-function and value function, starting with the last time step. These functions are given by
\begin{small}
\begin{align*}
V(\st) &= \frac{1}{2}\st\tr\Vxxt\st + \st\tr\Vxt + \text{const} \\
Q(\st,\at) &= \frac{1}{2}[\st;\! \at]\tr\Qyyt[\st;\! \at] \!+\! [\st;\! \at]\tr \Qyt \!+\! \text{const}
\end{align*}
\end{small}
We can express them with the following recurrence:
\begin{align*}
\Qyyt &= \costhesst + \fyt\tr\Vxxtp\fyt \\
\Qyt &= \costgradt + \fyt\tr\Vxtp \\
\Vxxt &= \Qxxt - \Quxt\tr\Quut\inv\Quxt \\
\Vxt &= \Qxt - \Quxt\tr\Quut\inv\Qut,
\end{align*}
\noindent which allows us to compute the optimal control law as \mbox{$\detpolicy(\state_t) = \haction_t + \kpol_t + \Kpol_t(\state_t - \hstate_t)$}, where $\Kpol_t = -\Quut\inv \Quxt$ and $\kpol_t = -\Quut\inv \Qut$. Performing a forward rollout using this control law allows us to find a new nominal trajectory, and the backward dynamic programming pass is repeated around this trajectory to take the next Gauss-Newton step.

To adapt DDP for performing MPC, we simply run the algorithm for a shorter horizon $H$ at each time time step $t$, so that the backward pass is performed from time step $t$ to $t + H$. 

\subsection{Adapting MPC for Guided Policy Search}
\label{sec:mpcgps}

While we could simply adapt the DDP-based MPC algorithm in the previous section to optimize Equation~\ref{eqn:trajobj}, the short horizon typically used in MPC makes it difficult to accomplish complex tasks like obstacle avoidance, which require long-horizon lookahead, with only a high-level specification of the task, such as a desired flight direction and an obstacle collision penalty. Instead, we use an offline optimization based on iterative LQG \cite{lt-ilqr-04} to first generate a reference trajectory, and then track this trajectory using MPC, with an additional term to account for differences from the neural network policy $\policy_\params(\at|\ot)$.

Offline, we run iterative LQG with our known approximate model to optimize Equation~\ref{eqn:trajobj}. Since Equation~\ref{eqn:trajobj} contains a KL-divergence term, the objective can be rewritten as
\begin{small}
\begin{multline}
\min_{\trajdist_i(\traj)} \sum_{t=1}^T E_{\trajdist_i(\st,\at)} \Big[ \frac{1}{\admmrho_t^i} \cost(\st,\at) - \frac{1}{\admmrho_t^i} \at\tr\lgmut^i \\
- \log \policy_\params(\at|\st) -
\vphantom{\frac{1}{\admmrho_t^i}}  \ent\big(\trajdist_i(\at|\st)\big) \Big], 
\label{eqn:offlineobj}
\end{multline}
\end{small}
\noindent where $\ent$ represents the maximum entropy, and this maximum entropy objective can be optimized with iterative LQG. The solution is a linear-Gaussian controller of the form \mbox{$\trajdist_i(\at|\st) = \gauss(\Kpol_t\st + \kpol_t,\Quut\inv)$}, as shown in prior work \cite{lk-gps-13}. Prior methods sample directly from the linear-Gaussian controller $\trajdist_i(\at|\st)$ \cite{lk-lcnnp-14}, but since we would like to use MPC to robustly control the robot during the rollout, we instead construct a surrogate cost function $\tilde{\cost}(\st,\at)$ for MPC that will allow us to robustly generate trajectories that have high probability under $\trajdist_i(\traj)$.

This surrogate cost should fulfill a number of criteria in order to be effective: first, it must encourage MPC to visit states that have high probability under $\trajdist_i(\traj)$; second, it must produce good long-horizon behavior even when optimized under a short horizon; and third, it must keep the generated behavior close to the neural network policy $\policy_\params(\at|\st)$. When we run MPC at time step $t$ from the current state $\st$, we first compute $\trajdist_i(\state_{t\pr}|\st)$ for each $t\pr\in [t+1,t+H]$ under the known approximate dynamics and the time-varying linear-Gaussian controller obtained from the offline LQG optimization. Using $\mu_{t\pr}$ and $\Sigma_{t\pr}$ to denote the mean and covariance of $\trajdist_i(\state_{t\pr}|\st)$, we can compute these distributions according to the following recurrence:
\begin{small}
\begin{align*}
\mu_{t\pr+1} &=
\left[ \begin{matrix}\fxtpr \hspace{-0.08in} &\futpr\end{matrix} \right] \!\!
\left[ \!\!
\begin{array}{l l}
\mu_{t\pr}\\
\hat{\action}_{t\pr} + \kpol_{t\pr} + \Kpol_{t\pr}(\mu_{t\pr} - \hat{\state}_{t\pr})
\end{array}
\!\! \right]\\
\Sigma_{t\pr+1} &=
\left[ \begin{matrix}\fxtpr \hspace{-0.08in} &\futpr\end{matrix} \right] \!\!
\left[ \!\!
\begin{array}{l l}
\Sigma_{t\pr} & \!\!\! \Sigma_{t\pr}\Kpol_{t\pr}\tr\\
\Kpol_{t\pr}\Sigma_{t\pr} & \!\!\! \Quutpr\inv \!+\! \Kpol_{t\pr}\Sigma_{t\pr}\Kpol_{t\pr}\tr
\end{array}
\!\! \right] \!\!
\left[ \begin{matrix}\fxtpr\tr \\ \futpr\tr\end{matrix} \right] \!+\! \noise_{t\pr}
\end{align*}
\end{small}
The intuition here is that we would like to figure out which states the offline LQG solution would prefer to visit, independently of the actions required to get to these states. This is important since, in the presence of model errors and perturbations, the nonlinear approximate model might indicate different actions when combined with MPC, but the overall distribution over states should remain similar. Once we obtain $\trajdist_i(\state_{t+1},\dots,\state_{t+H}|\st)$, we can marginalize to obtain $\trajdist_i(\state_{t\pr})$ for each time step $t\pr \in [t + 1, t + H]$, and we then construct the surrogate cost as
\begin{multline}
\tilde{\cost}(\state_{t\pr},\action_{t\pr}) = -\log \trajdist_i(\state_{t\pr}|\st)\\
- \admmrho^i_{t\pr}\log \policy_\params(\action_{t\pr}|\state_{t\pr}) - \action_{t\pr}\tr\lgmutpr^i. \label{eqn:mpcobj}
\end{multline}
We then run MPC on this cost as described in the previous section to obtain a new linear-Gaussian controller for time step $t$ of the form \mbox{$\tilde{\trajdist}_{ij}(\at|\st) = \gauss(\tilde{\Kpol}_{tij}\st + \tilde{\kpol}_{tij},\tQuutij\inv)$}, and choose the action by sampling from this linear-Gaussian. The subscript $ij$ here denotes the $j\nth$ sample (generated via MPC) from the $i\nth$ trajectory distribution.

While samples generated in this way are not exactly samples from $\trajdist_i(\traj)$, but rather samples from a distribution formed by the product of independent marginals at each time step, we found that the resulting algorithm was still able to produce good training data for the neural network in guided policy search. Furthermore, the additional information provided by $\trajdist_i(\state_{t\pr}|\st)$ allowed MPC to succeed in the presence of model errors and disturbances, even on tasks such as obstacle avoidance that require long-horizon lookahead.

One final detail is that both the cost in Equation~(\ref{eqn:mpcobj}) and the offline optimization objective in Equation~(\ref{eqn:offlineobj}) require access to $\log \policy_\params(\at|\st)$, while we only have access to $\log \policy_\params(\at|\ot)$, and $\ot$ is in general a complex and unknown function of $\st$, since the observation might include, e.g., laser rangefinder readings, while the state might consist of the vehicle's position and orientation. To obtain $\log \policy_\params(\at|\st)$, we follow prior work and approximately linearize the policy by using the previous set of rollouts from the physical system. This can be done by fitting a time-varying linear-Gaussian model of $\log \policy_\params(\at|\st)$ to the samples, since each sample includes both $\st$ and $\ot$, allowing us to evaluate the policy at each sampled state $\st$. The fitting is done by using linear regression with a Gaussian mixture model prior, as in previous work \cite{la-lnnpg-14}.

\begin{algorithm}[b]
\caption{MPC-guided policy search}
\label{alg:mpcgps}
\begin{algorithmic}[1]
\FOR{iteration $k=1$ to $K$}
\STATE Optimize $\trajdist_i(\traj)$ offline according to Equation~(\ref{eqn:offlineobj})
\STATE Run MPC $M$ times from initial states $\state_1 \sim \trajdist_i(\state_1)$ to create $\{\tilde{\trajdist}_{ij}(\traj)\}$ and $\{\traj_{ij}\}$ using $\tilde{\cost}(\traj)$ in Equation~(\ref{eqn:mpcobj})
\STATE Train nonlinear policy $\policy_\params(\at|\ot)$ to match each $\tilde{\trajdist}_{ij}(\at|\st)$ along each $\traj_{ij}$, using Equation~(\ref{eqn:polobj})
\STATE Fit time-varying linear-Gaussian model to estimate $\policy_\params(\at|\st)$ around each $\trajdist_i(\traj)$ using samples $\{\traj_{ij}\}$
\STATE Update $\admmrho_t^i$ and $\lgmut^i$ as in~\cite{levine2015end}
\ENDFOR
\STATE {\bf return} optimized policy parameters $\params$
\end{algorithmic}
\end{algorithm}

\subsection{Training the Nonlinear Policy}

The final nonlinear policy $\policy_\params(\at|\ot)$ is trained using standard supervised learning, from samples collected via MPC. The objective for this supervised learning is given in Equation~(\ref{eqn:polobj}), though in the case of MPC-based samples, we substitute $\tilde{\trajdist}_{ij}(\at|\st)$ for $\trajdist_i(\at|\st)$. In the case of a conditionally Gaussian policy $\policy_\params(\at|\ot) = \gauss(\polmu(\ot),\polsig(\ot))$, the KL-divergence $\kl(\policy_\params(\at|\ot^{i,j})\|\tilde{\trajdist}_{ij}(\at|\st^{i,j}))$ in this objective can be written out as
\begin{small}
\begin{multline*}
\kl\big(\policy_\params(\at|\ot^{i,j})\|\tilde{\trajdist}_{ij}(\at|\st^{i,j})\big) = \\
\frac{1}{2}\big(\polmu(\ot) - \tilde{g}_{tij}(\st)\big)\tQuutij\big(\polmu(\ot) - \tilde{g}_{tij}(\st)\big) \\
- \frac{1}{2}\trace\big[\tQuutij\polsig(\ot)\big] + \frac{1}{2}\log|\polsig(\ot)| + \lgmut\tr\polmu(\ot),
\end{multline*}
\end{small}%
where $\tilde{g}_{tij}(\st) = \tilde{\Kpol}_{tij} \st + \tilde{\kpol}_{tij}$. Note that this is simply a weighted least squares objective on the mean function $\polmu(\ot)$. In this work, we represent $\polmu(\ot)$ with a multilayer neural network, which allows us to train flexible and expressive policies. Since we prefer deterministic or nearly-deterministic policies, we choose $\polsig(\ot)$ to be constant, which means that we can solve for it in closed form according to
\begin{small}
\begin{equation*}
\polsig(\ot) = \left(\frac{1}{N}\sum_{i=1}^N \sum_{j=1}^M \tQuutij\right)\inv.
\end{equation*}
\end{small}%
The neural network mean function $\polmu(\ot)$ is optimized using stochastic gradient descent (SGD). As noted earlier, one of the key advantages of this type of training approach is that the input $\ot$ to the neural network policy need not match the state $\st$ used during trajectory optimization and MPC, which allows us to train policies that operate directly on raw inputs from the onboard sensors.

\subsection{Algorithm Summary}

A summary of our method is presented in Algorithm~\ref{alg:mpcgps}. At each iteration, we first generate an offline solution by using iterative LQG to optimize the augmented objective in Equation~(\ref{eqn:offlineobj}). This offline solution allows us to initialize and construct the cost for MPC rollouts. We conduct $M$ MPC rollouts for each trajectory distribution $\trajdist_i(\traj)$, constructing a new surrogate cost $\tilde{\cost}(\st,\at)$ at each time step. These MPC rollouts provide us with sample trajectories $\{\traj_{ij}\}$ and MPC controllers $\{\tilde{\trajdist}_{ij}(\traj)\}$, which we can use to train the nonlinear neural network policy $\policy_\params(\at|\ot)$ as described in the previous section. After the policy is trained, we update our time-varying linear-Gaussian fit for $\policy_\params(\at|\st)$ by using the latest samples. Note that a separate linear-Gaussian estimate of $\policy_\params(\at|\st)$ is constructed around each trajectory distribution $\trajdist_i(\traj)$. Finally, we adjust the dual variables as described in previous work \cite{levine2015end}.

\section{Experimental Evaluation}

\newcommand{\specialcell}[2][c]{%
  \begin{tabular}[#1]{@{}c@{}}#2\end{tabular}}
\renewcommand{\tabcolsep}{0pt}

\begin{table*}
\vspace*{5pt}
\begin{center}
\footnotesize{
\begin{tabular}{|C{0.12\linewidth}*{3}{|C{0.07\linewidth}}|*{3}{|C{0.07\linewidth}}|*{3}{|C{0.07\linewidth}}|*{3}{|C{0.07\linewidth}}|}
\hline
& \multicolumn{12}{c|}{train: single cylinder; test: infinite forest} \\ \hline
&  \multicolumn{3}{c||}{no model error} &  \multicolumn{3}{c||}{0.05kg mass error} & \multicolumn{3}{c||}{$8\%$ rotor bias} & \multicolumn{3}{c|}{perturbed model parameters} \\ \hline
method & \specialcell{(baseline)\\offline\\$\cost(\st,\at)$} & \specialcell{(baseline)\\MPC\\$\cost(\st,\at)$} & \specialcell{{\bf full}\\MPC\\$\tilde{\cost}(\st,\at)$} & \specialcell{(baseline)\\offline\\$\cost(\st,\at)$} & \specialcell{(baseline)\\MPC\\$\cost(\st,\at)$} & \specialcell{\specialcell{\bf full}\\MPC\\$\tilde{\cost}(\st,\at)$} & \specialcell{(baseline)\\offline\\$\cost(\st,\at)$} & \specialcell{(baseline)\\MPC\\$\cost(\st,\at)$} & \specialcell{{\bf full}\\MPC\\$\tilde{\cost}(\st,\at)$} & \specialcell{(baseline)\\offline\\$\cost(\st,\at)$} & \specialcell{(baseline)\\MPC\\$\cost(\st,\at)$} & \specialcell{\specialcell{\bf full}\\MPC\\$\tilde{\cost}(\st,\at)$} \\ \hline
\specialcell{number of \\ training crashes} &
1 & 0 & {\bf 0} & 
1 & 0 & {\bf 0} &
46 & 0 & {\bf 0} & 
N/A & 4 & {\bf 0} \\
\hline
\specialcell{average test \\ flight duration (s)} & 
\specialcell{{\bf 56.9} \\$\pm$ 27.3} & 
\specialcell{35.8 \\$\pm$ 22.5} & 
\specialcell{53.4 \\$\pm$ 22.5} & 
\specialcell{11.3 \\$\pm$ 5.2} & 
\specialcell{6.9  \\$\pm$ 2.6} & 
\specialcell{{\bf 34.4} \\$\pm$ 8.4} & 
\specialcell{60.0 \\$\pm$ 15.8} & 
\specialcell{56.1 \\$\pm$ 30.8} & 
\specialcell{{\bf 95.7} \\$\pm$ 9.8} & 
\specialcell{N/A} & 
\specialcell{20.8 \\$\pm$ 9.8} & 
\specialcell{{\bf 60.5} \\$\pm$ 27.0} \\
\hline
\end{tabular}
}
\end{center}
\begin{center}
\footnotesize{
\begin{tabular}{|C{0.12\linewidth}*{3}{|C{0.07\linewidth}}|*{3}{|C{0.07\linewidth}}|*{3}{|C{0.07\linewidth}}|*{3}{|C{0.07\linewidth}}|}
\hline
& \multicolumn{12}{c|}{train: straight hallway; test: winding hallway} \\ \hline
&  \multicolumn{3}{c||}{no model error} &  \multicolumn{3}{c||}{0.05kg mass error} & \multicolumn{3}{c||}{$8\%$ rotor bias} & \multicolumn{3}{c|}{perturbed model parameters} \\ \hline
method & \specialcell{(baseline)\\offline\\$\cost(\st,\at)$} & \specialcell{(baseline)\\MPC\\$\cost(\st,\at)$} & \specialcell{{\bf full}\\MPC\\$\tilde{\cost}(\st,\at)$} & \specialcell{(baseline)\\offline\\$\cost(\st,\at)$} & \specialcell{(baseline)\\MPC\\$\cost(\st,\at)$} & \specialcell{\specialcell{\bf full}\\MPC\\$\tilde{\cost}(\st,\at)$} & \specialcell{(baseline)\\offline\\$\cost(\st,\at)$} & \specialcell{(baseline)\\MPC\\$\cost(\st,\at)$} & \specialcell{{\bf full}\\MPC\\$\tilde{\cost}(\st,\at)$} & \specialcell{(baseline)\\offline\\$\cost(\st,\at)$} & \specialcell{(baseline)\\MPC\\$\cost(\st,\at)$} & \specialcell{\specialcell{\bf full}\\MPC\\$\tilde{\cost}(\st,\at)$} \\ \hline
\specialcell{number of \\ training crashes} &
0 & 0 & {\bf 0} & 
0 & 0 & {\bf 0} &
76 & 0 & {\bf 0} & 
N/A & 0 & {\bf 0} \\
\hline
\specialcell{average test \\ flight duration (s)} & 
\specialcell{{\bf 46.2} \\$\pm$ 28.4} & 
\specialcell{35.2 \\$\pm$ 13.3} & 
\specialcell{35.2 \\$\pm$ 20.0} & 
\specialcell{21.7 \\$\pm$ 5.8} & 
\specialcell{15.7  \\$\pm$ 1.3} & 
\specialcell{{\bf 31.8} \\$\pm$ 15.7} & 
\specialcell{26.0 \\$\pm$ 21.1} & 
\specialcell{{\bf 51.1} \\$\pm$ 28.6} & 
\specialcell{28.5 \\$\pm$ 16.2} & 
\specialcell{N/A} & 
\specialcell{9.9 \\$\pm$ 4.2} & 
\specialcell{{\bf 55.2} \\$\pm$ 17.5} \\
\hline
\end{tabular}
}
\caption{
Training and test results comparing our full MPC-guided policy search method with two baselines. The test flight duration was averaged over 20 runs of the learned policies.
In one scenario (top table), the policy search method variants were trained on a single cylinder and tested in an infinite forest. In the other scenario (bottom table), the policy search variants were trained in a straight hallway and tested in a winding hallway. Each policy search variant was tested with four different model errors. Experiments labelled N/A were unable to complete due to excessive crashing.
In the majority of experiments, the full MPC-guided policy search method outperformed the two baselines, crashing less during training and with the final learned neural network policy flying for the longest duration in the test scenarios.
\label{tbl:generalization}
}
\end{center}
\vspace*{-10pt}
\end{table*}

We evaluated our method on simulated quadrotor obstacle avoidance tasks.

\subsection{Quadrotor System}

The simulated quadrotor was modeled after 3DR's IRIS+\footnote{\url{https://3dr.com/kb/iris/}}, which has width 0.47m, height 0.11m, and weight 1.5kg. The dynamics followed the formulation described by Martin and Salaun \cite{ms-trafq-10}. The true state of the vehicle ${\st=(\stp, \stv, \stq, \stw) \in \mathbb{R}^{13}}$ consisted of the position ${\stp=(x_t, y_t, z_t)}$ and orientation $\stq$, expressed as a quaternion, as well as their time derivatives, i.e. linear velocity $\stv$ and angular velocity $\stw$. The controls ${\at \in \mathbb{R}^4}$ consisted of rotor velocities. The observation model ${\ot=(\str, \stv, \stq, \stw) \in \mathbb{R}^{40}}$ lacked the position $\stp$ and instead included readings $\str$ from a set of 30 equally spaced laser rangefinders with max range 5m arranged in 180 degree fan in front of the vehicle. This type of observation model is quite challenging to integrate into simple control methods, such as time-varying linear controllers, but can be easily processed by a multilayer neural network policy.

\subsection{Cost Function}

The cost function for the iterative offline LQG optimization was
\begin{small}
\begin{align*}
l(\st, \at) = 
&10^3 ||\stv - \mathbf{v}^*||_2^2 + 
 500  ||z_t - z^*||_2^2 + \\
&10^4 ||\stq - \mathbf{q}^*||_2^2 +
 250  ||\stw - \boldsymbol{\omega}^*||_2^2 + \\
&0.5||\at - \bu_{\textsc{hover}}^*||_2^2 + \\
&10^4 \max(d_{\textsc{safe}} - \text{signed\_distance}(\st), 0)
\end{align*}
\end{small}%
where $\st=((x_t, y_t, z_t), \stv, \stq, \stw)$ is the full state as defined previously;
$z^*, \mathbf{v}^*, \mathbf{q}^*, \boldsymbol{\omega}^*$ are the target height, linear velocity, orientation, and angular velocity, respectively; and $\bu_{\textsc{hover}}^*$ is the desired rotor velocity for hovering. The final term is a hinge loss on the distance of the quadrotor to the nearest obstacle; there is no penalty if the nearest obstacle is further than $d_{\textsc{safe}}$.

\begin{figure}
\vspace*{10pt}
\centering
\includegraphics[height=0.32\columnwidth]{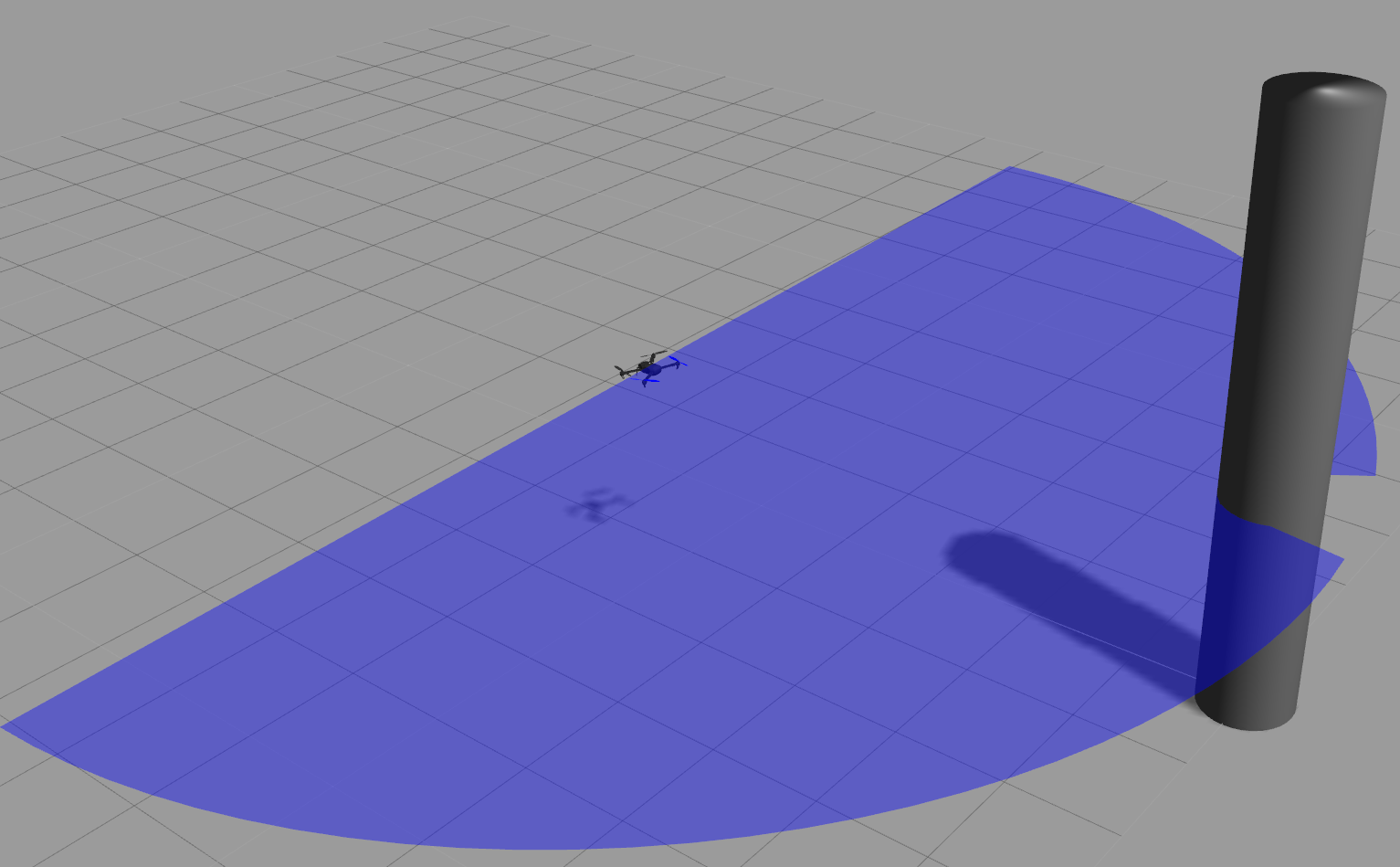}
\includegraphics[height=0.32\columnwidth]{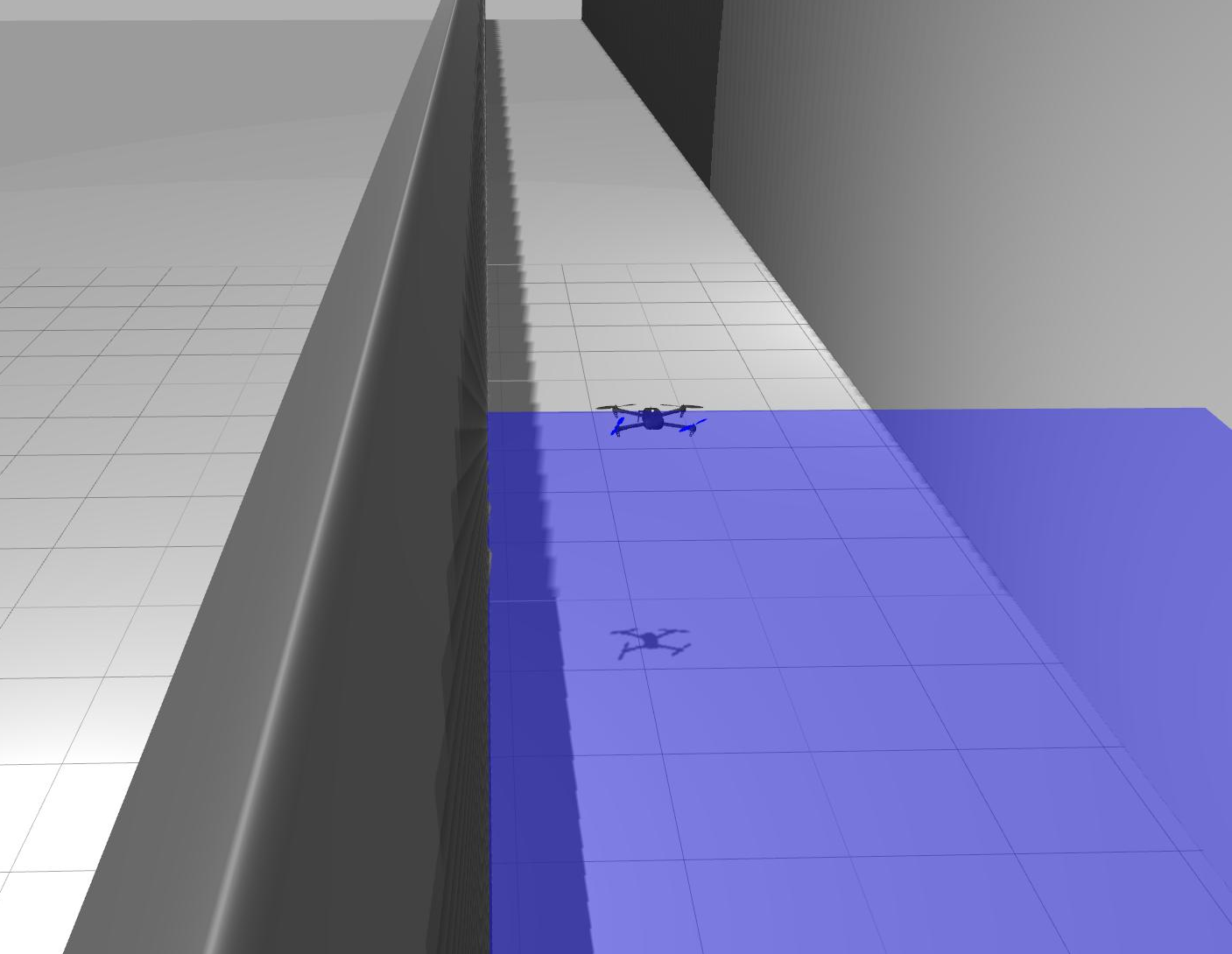}
\caption{The quadrotor must learn to fly around a cylindrical obstacle and down a hallway using only onboard sensing. The blue semicircle is the range of the onboard laser range finders.}
\label{fig:scenarios}
\end{figure}

\subsection{Neural Network Policy}

The neural network policy consisted of two fully connected hidden layers, each with 40 rectified linear (ReLU) units \cite{Nair2010_ICML}. For training the neural network in each iteration of guided policy search, we used the ADAM \cite{Kingma2015_ICLR} algorithm, optimizing using 20,000 minibatches with a minibatch size of 50. We used default values for all learning parameters as presented in \cite{Kingma2015_ICLR}.

%


\subsection{Experimental Domains}

Figure~\ref{fig:scenarios} depicts the two simulated environments in which we trained our neural network policies: a single cylindrical obstacle of radius 0.5m and height 4m, and a straight hallway of width 5m and height 4m. For the cylinder avoidance task, we used $N=18$ initial states with varying initial $(x, y)$ positions in front of the cylinder; for the hallway task, we used $N=6$ initial states with varying $y$-coordinate values. For both tasks, each initial state corresponds to a different trajectory distribution $\trajdist_i(\traj)$, with $M=4$ samples from each distribution. Each trajectory $\traj$ has length $T=150$, which equates to 7.5 seconds. Collision with an obstacle or the ground, or flying above an obstacle were considered as a training crash. 

\subsection{Baseline Methods and Model Errors}

To evaluate the importance of using MPC, we trained neural network policies on each of the tasks in the presence of model errors, using three variants: the full MPC-guided policy search algorithm with the surrogate cost $\tilde{\cost}(\st,\at)$ described in Section~\ref{sec:mpcgps}\;, a variant of MPC-guided policy search that uses the true cost $\cost(\st,\at)$ with the policy KL-divergence term and dual variables for MPC, and a variant that does not use MPC at all, and instead performs the rollouts by using the time-varying linear-Gaussian policy generated by the offline iterative LQG algorithm. All methods were trained for $K=5$ guided policy search iterations.

We evaluated each of the above variants in the presence of four different types of model errors: no model error, the actual weight of the quadrotor was 0.05kg (3.3\%) greater than the expected weight, the two rotors on one side of the quadrotor had an 8\% multiplicative rotor velocity bias, and all model parameters (e.g. moments of inertia, drag coefficients) were perturbed by retaining only one significant digit.

\subsection{Results}

Table~\ref{tbl:generalization} shows the number of crashes experienced by the quadrotor during training. These results indicate that MPC using the surrogate cost is able to train a successful neural network policy without experiencing catastrophic failure.


To evaluate the generalization of the learned policies, we ran the trained neural networks in two test scenarios: an infinite forest of cylinders of the same shape as in training, but at random positions an average distance of 5m apart, and a winding hallway with randomized turns of at most $30^{\circ}$ every 5m. We ran the policies trained with a single cylinder in the infinite forest, and we ran the trained policies for the straight hallway in the winding hallway. The average flight duration of the final trained neural network policies are shown in Table~\ref{tbl:generalization}. With no model errors, our MPC-guided policy search algorithm was comparable to the other methods. When model errors were introduced, our method outperformed the two baselines in the majority of scenarios. Videos of the resulting flights are included as supplementary material, and may also be viewed on the project webpage\footnote{\url{http://rll.berkeley.edu/icra2016mpcgps}}.

Our evaluation shows that MPC-guided policy search is an effective algorithm for off-policy training of complex neural network policies for autonomous aerial vehicles. Our full method was able to learn each of the two behaviors without experiencing any catastrophic failures during training, and the trained policy was able to generalize effectively.

\section{Discussion and Future Work}

We presented an algorithm for training deep neural network control policies for autonomous aerial vehicles, by using model predictive control to generate guiding samples for guided policy search. Our MPC-guided policy search uses a modified MPC algorithm that trades off minimizing the cost against matching the current neural network policy, so as to generate good training data that can be used to train a better policy with standard supervised learning. Since the partially trained neural network policy is never used to choose actions at training time, the more robust and reliable MPC method provides a substantial improvement in safety over traditional reinforcement learning methods. Our results show that this algorithm is able to learn complex policies, such as high speed obstacle avoidance, using raw sensor inputs and low-level rotor command outputs.

One of the key ideas behind our method is the notion of an instrumented training setup, which allows MPC to be performed at training time with full state observations, which could be provided, for example, by using motion capture. At the same time, the vehicle gathers observations from its own onboard sensors, and trains the policy to mimic the action chosen by MPC using only the raw sensor readings, without relying on the full state. Acquiring the sensor readings is important, because accurately modeling complex sensors, such as laser range finders and cameras, is very difficult, while obtaining a model of the vehicle that is accurate enough to perform MPC is comparatively easier.

While our approach can train very complex, high-dimensional policies, it shares many of the limitations of prior guided policy search methods \cite{levine2015end}. In particular, full state observations are required at training time, in order to perform MPC, even though the final neural network policy can perform the task using only onboard sensors. In the real world, this kind of state information could be obtained using an instrumented training environment (with, for example, motion capture). Since the instrumentation is only required during training, the final neural network is still able to act in the real world, so this approach is practical for a wide range of robotic tasks. However, not all aerial maneuvers can be learned in such an instrumented training setup, and combining explicit state estimation with guided policy search in future work could lead to a much more broadly applicable algorithm. Another direction that can be explored in future work is to combine guided policy search with more sophisticated MPC and planning algorithms. In principle, a wide variety of methods can be used to generate guiding samples, and more sophisticated methods might afford superior robustness and obstacle avoidance \cite{dt-emip-15}.



\section*{Acknowledgements}

This research was funded in part by the Army Research Office through the MAST program, by DARPA under Award \#N66001-15- 2-4047, and by the Berkeley Vision and Learning Center (BVLC).

\bibliographystyle{ieeetran}
\bibliography{references}

\end{document}